# Blind estimation of white Gaussian noise variance in highly textured images

*Mykola Ponomarenko[a], Nikolay Gapon[b], Viacheslav Voronin[b], Karen Egiazarian[a]*
[a] *Tampere University of Technology, FIN 33101, Tampere, Finland;*
[b] *Don State Technical university, 344000, Rostov-on-Don, Russian Federation*

## Abstract

*In the paper, a new method of blind estimation of noise variance in a single highly textured image is proposed. An input image is divided into 8x8 blocks and discrete cosine transform (DCT) is performed for each block. A part of 64 DCT coefficients with lowest energy calculated through all blocks is selected for further analysis. For the DCT coefficients, a robust estimate of noise variance is calculated. Corresponding to the obtained estimate, a part of blocks having very large values of local variance calculated only for the selected DCT coefficients are excluded from the further analysis. These two steps (estimation of noise variance and exclusion of blocks) are iteratively repeated three times. For the verification of the proposed method, a new noise-free test image database TAMPERE17 consisting of many highly textured images is designed. It is shown for this database and different values of noise variance from the set {25, 49, 100, 225}, that the proposed method provides approximately two times lower estimation root mean square error than other methods.*

*Keywords: Blind estimation of noise characteristics, discrete cosine transform (DCT), noise free test image database*

## Introduction

Task of blind estimation of noise variance takes a prominent place in digital image processing [1-7]. Knowing of noise variance is required for solving of wide range of problems such as image denoising, lossy image compression (automatic selection of an optimal operation point), no reference assessment of image visual quality, digital watermarking, etc [5-9].

In particular, to effectively perform an image denoising, one should be able to estimate noise variance with an error lower than 20% from the true value of the variance [10-12]. Under-estimated or over-estimated values of variance may result in a non-efficient filtering (noise is present in filtered images) or in over-smoothing of images (image details are smoothed), respectively.

Nowadays, there exist many methods of blind noise estimation. Some of them deal with robust estimates in spatial or spectral domain [13-19]. To decrease influence of texture, most of methods perform preliminary detection of homogeneous regions using segmentation, histogram based and other approaches [20-28]. There are non-local methods used to estimate noise variance on the differences between similar image regions [29] as well as methods estimating noise variance in fractal domain (analyzing self-similarity of image regions in different scales) [30], and methods based on principal components analysis [31].

Because for many practical cases noise in images is spatially correlated, there are methods intended for estimations of variance of spatially correlated noise [32] as well as its spectrum [33].

Often estimated noise is multiplicative or non-Gaussian. For example, noise may have Poisson or Rayleigh distribution. There are methods for estimation of characteristics of such multiplicative or non-Gaussian noises [34-40], mixed additive and multiplicative noise [39]. One should note that in some cases, a problem of estimation of parameters of multiplicative noise may by replaced by estimation of variance of additive noise by preliminary usage to analyzed images of homomorphic or variance stabilizing transforms [41-43], such as Anscombe transform [44].

Finally, there are methods for blind estimation of noise parameters for multichannel images, which take into account the presence of correlation between pixels of neighbor channels [42-49].

Despite the fact that the problem of blind noise parameters estimation is well studied, even best methods not always provide desirable precision of obtained estimates.

For images with large homogeneous areas, most of the methods of blind noise variance estimation are able to produce accurate enough estimates. However, in practice, there exist many images with no homogeneous regions. For such images, even most effective from existing methods provides over-estimated (often in several times) estimates of noise variance.

The main goal of this paper is a design of a new method able to estimate noise variance on a highly textured image with an accuracy appropriate for image denoising. In this respect, a new noise free test image database containing many highly textured images is presented in the paper. An existence of such a noise free test image database is necessary for the verification of methods of blind noise variance estimation.

## Description of the proposed method

The proposed method is based on an assumption that analyzed noise is additive white Gaussian noise. For a spatially correlated noise it is unusable. For a multiplicative noise, usage of the method is possible after application of the homomorphic transform corresponding to the noise distribution.

An analyzed image is divided into N blocks of 8x8 pixels with an overlap. For each block, the two-dimensional DCT is performed. As a result, there are 64 DCT coefficients in N blocks (matrix 64xN).

Let $D(i,j)$ denotes a value of $j$-th DCT coefficient of $i$-th block, where $j = 1…64$, $i = 1…N$. For each $j$ sum $E(j)$ of squares of $D(i,j)$ is calculated:

$$E(j) = \sum_{i=1}^{N} D(i,j)^2 . \qquad (1)$$

Thus, $E(j)$ is a measure of energy of $j$-th DCT coefficient for all analyzed blocks of the image.

K lower values of $E(j)$ are selected. K corresponding DCT coefficients from each block are grouped into a column vector. These column vectors for all N blocks are grouped into a 2D matrix **A** for a further analysis (see Fig. 1).

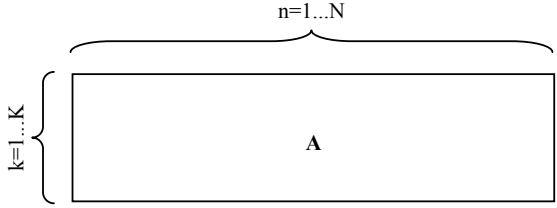

Fig. 1. 2D matrix **A** used for noise variance estimation

The remaining 64 – K DCT coefficients of each image block are excluded from the further analysis.

The parameter K is selected from the range of 4 to 60. Lower value of K provides better robustness to a presence of textures. Higher value of K provides better accuracy of noise variance estimates. In the paper we set K=49.

Let $m$ be the index of row in **A** which corresponds to the lowest E(j) value $E_{min}$. If E(64) < 1.3 $E_{min}$ then $m$ is replaced by index of row in **A** corresponding to E(64). (Factor 1.3 here is obtained empirically).

A robust estimate S of noise standard deviation is calculated using median absolute deviation measure of $m$-th row of **A**:

$$z = \text{median}(\{A(m,1), A(m,2), ..., A(m,N)\}),$$
$$S = 1.4826 \, \text{median}(|A(m,1)-z|, |A(m,2)-z|, ..., |A(m,N)-z|). \quad (2)$$

Then the following four steps are repeated three times:

1. For each column of **A**, an estimate of local variance LD(n) is calculated as:

$$LD(n) = \frac{1}{K} \sum_{k=1}^{K} A(k,n)^2 . \quad (3)$$

2. A threshold $T = S^2(1+(8/K)^{0.5})$ for LD(n) values is calculated.

3. Columns of **A** with LD(n) larger than T are rejected from the matrix **A** and excluded from the further analysis because these columns correspond to image blocks of non-homogeneous regions. The value of N is decreased as a result of this step.

4. If N is larger than 1024 (this value is obtained empirically) then the new value of S is calculated according to (2).

After repeating three time of the steps 1-4, S contains obtained estimate of a standard deviation of noise. The final estimate of a noise variance is calculated as $S^2$.

Note that the proposed method combines an iterative calculation of estimates of noise variance in one row of the matrix **A** (along N-th dimension) excluding non-homogeneous blocks using local variances calculated in the columns of the matrix (along the K-th dimension). It allows to effectively exclude an informative component from the analysis without noticeable under-estimation of a noise variance.

Matlab code of the proposed method is available in http://ponomarenko.info/iedd.html.

## Noise-free test image database

For the verification of the proposed method, a noise-free test image database TAMPERE17 is designed. We had a following motivation for the image database creation. It is well known, that for a verification of many methods of digital image processing, e.g. image denoising, lossy compression, super resolution, blind noise and blur parameters estimation, researchers should use noise-free images. Moreover, it is very desirable to have images which are not lossy compressed or interpolated. However, there are no large image databases satisfying these conditions. Researchers have to use such test image database as TID2008 [50], Kodak, ALOT [51] (reference images have noise with unknown variance), NED2012 (reference images have noise with estimated variance) [30], BSD (reference images are lossy compressed using JPEG standard) [52].

TAMPERE17 database has 300 test images of 512x512 pixels (both color and grayscale versions). Images for the database are obtained with monitoring of noise level (the photo shooting was carried out on calibrated cameras with fixed ISO) and without any interpolation, lossy compression or other spatial processing (denoising, sharpening, etc.). For each image, a fragment 3072x3072 pixels or 2560x2560 pixels (depending on digital camera) was selected, and downscaled to 512x512 (by averaging of pixel values for the regions of 6x6 pixels or 5x5 pixels, respectively). As a result, only images with standard deviation of noise less than 1 have been included in the database. For the grayscale version of images, the noise level is even lower (values of noise standard deviation are in the text file included into the database).

The images for TAMPERE17 have been selected to present different kinds of images: highly textured images, images with large homogeneous regions, images with self-similar textures, images where only part of the image is in focus, etc (see Fig. 2).

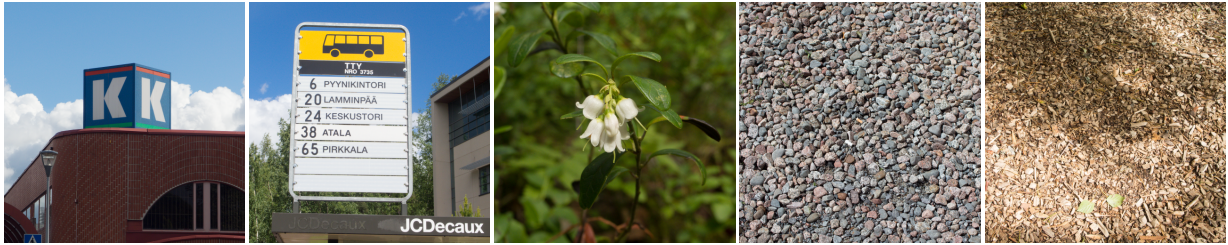

Fig. 2. Examples of noise free test images from TAMPERE17 database

TAMPERE17 database is freely available in http://www.cs.tut.fi/sgn/imaging/tampere17.

## Comparative analysis

Comparative analysis of effectiveness of the proposed method is carried out for 4 values of noise variance: 25, 49, 100 and 225. All 300 grayscale images of TAMPERE17 were distorted by an additive white Gaussian noise with a given variance. After, the variance was estimated for each image by each of the compared methods.

For the comparison, the following methods have been used: the proposed method - iterative estimation in DCT domain (IEDD), method RDCT [29], method WTP [20] and method PCA [31].

Both relative RMSE (RRMSE) of variance estimates and number of abnormal estimates (AE) are used as quality criteria. Abnormal estimate is an estimate of noise variance which differs from the true value on more than 20%.

RRMSE is calculated in the paper as

$$\text{RRMSE} = 100\% \sqrt{\frac{1}{300} \sum_{i=1}^{300} \left( \frac{\sigma_{est}^2 - \sigma_{ref}^2}{\sigma_{ref}^2} \right)^2} . \quad (4)$$

Here $\sigma_{est}^2$ is the estimated noise variance, $\sigma_{ref}^2$ is the true noise variance.

For all variance values, the proposed method IEDD provides approximately 2 times lower RRMSE than other compared methods (see Fig. 3). Moreover, IEDD is the only method which for variance 25 provides RRMSE lower than 20% (as a required criterion for further effective noise suppression [10, 11]).

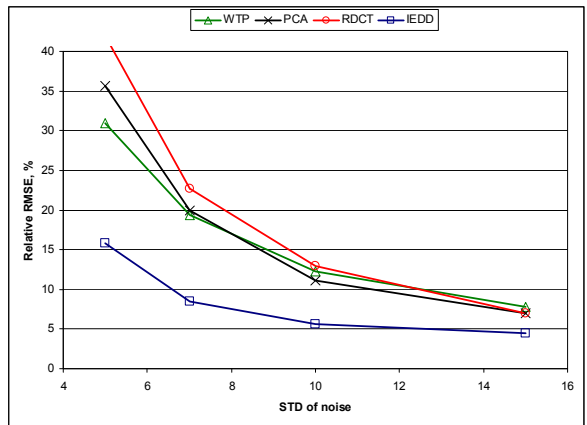

Fig. 3. Relative root mean square error of blind noise variance estimation for TAMPERE17 database

In correspondence with AE criterion (see Fig. 4), IEDD also significantly outperforms all compared methods.

For example, for variance 25 following values of AE are obtained: IEDD=23, RDCT=49, PCA=97, WTP=88. As it is seen, the number of abnormal estimates for IEDD are 2 times lower than that for nearest competitor and 4 times lower than for WTP method.

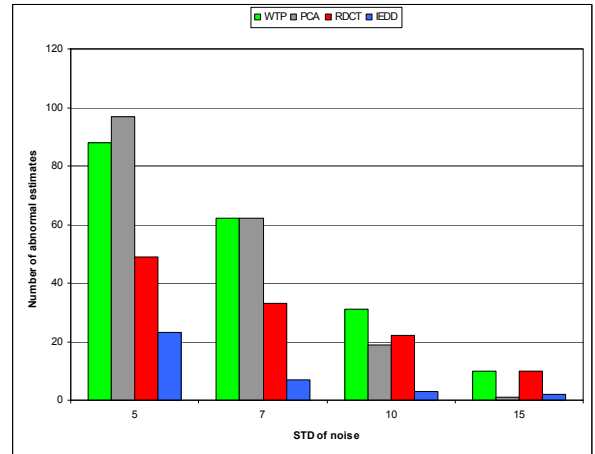

Fig. 4. Number of abnormal estimates (more than 20% difference between an estimate and true value of noise variance)

Computational complexities (characterized by computational time for processing a 512x512 grayscale image) of compared methods are shown in Table 1.

**Table 1. Computational time for compared methods**

| # | Method | Time, sec |
|---|--------|-----------|
| 1 | WTP    | 0.74      |
| 2 | IEDD   | 0.89      |
| 3 | PCA    | 1.57      |
| 4 | RDCT   | 1.98      |

In our experiments we use: 64-bit Windows 8.1, Intel(R) Core(TM) i7-4500U 2.4 GHz, 8 GB RAM, MATLAB 2016a.

All methods except of RDCT use only MATLAB code. RDCT is written using Delphi programming language.

As it seen from the Table 1, complexity of the proposed IEDD method is comparable with the fastest (in the comparison) WTP method. The most simple way to speed up the IEDD method is decreasing of N value (number of analyzed blocks of an image).

## Conclusions

The proposed method IEDD effectively combines analysis of frequency range in DCT domain with lower presence of informative component, iterative exclusion from the analysis the blocks with high local variance in the frequency range, and a robust estimation of noise variance by values of one DCT coefficients for all blocks. Both a link to Matlab source code of the IEDD method and a link to noise free test image database TAMPERE17 are given.

The proposed method IEDD provides twice lower error of blind variance of noise estimation than that of the nearest competitors providing state-of-the-art results for blind estimation of variance of additive white noise from a single highly textured image.

## Acknowledgment

This work is supported by Academy of Finland, project no. 287150, 2015-2019.